\documentclass[runningheads]{llncs}

\usepackage{eccv}

\usepackage{eccvabbrv}

\usepackage{graphicx}
\usepackage{booktabs}
\usepackage{amsmath}
\usepackage{adjustbox}

\def\thickerhline{\noalign{\hrule height1.5pt}}
\usepackage{tabularx}
\usepackage{multirow}
\usepackage{pifont}%
\newcommand{\cmark}{\ding{51}}%
\newcommand{\xmark}{\ding{55}}%

\usepackage[accsupp]{axessibility}  %

\usepackage[pagebackref,breaklinks,colorlinks,citecolor=eccvblue]{hyperref}

\usepackage{orcidlink}

\begin{document}

\title{Object Pose Estimation Using Implicit Representation For Transparent Objects} 

\titlerunning{Object Pose Estimation Using Implicit Representation}

\author{Varun Burde\thanks{Equal Contribution}\inst{1,2}\orcidlink{0000-0001-8317-6164}  \and
Artem Moroz$^\star$\inst{2}\orcidlink{0009-0007-5831-7106} \and
Vít Zeman$^\star$\inst{2}\orcidlink{0009-0007-9304-9354}
\and
Pavel Burget \inst{1,2}\orcidlink{0000-0002-4787-8182}}

\authorrunning{Burde et al.}

\institute{Czech Technical University in Prague,
Czech Republic \and
Czech Institute of Informatics, Robotics and Cybernetics, Czech Republic
}

\maketitle

\begin{abstract}

Object pose estimation is a prominent task in computer vision. The object pose gives the orientation and translation of the object in real-world space, which allows various applications such as manipulation, augmented reality, etc. Various objects exhibit different properties with light, such as reflections, absorption, etc. This makes it challenging to understand the object's structure in RGB and depth channels. Recent research has been moving toward learning-based methods, which provide a more flexible and generalizable approach to object pose estimation utilizing deep learning. One such approach is the render-and-compare method, which renders the object from multiple views and compares it against the given 2D image, which often requires an object representation in the form of a CAD model. We reason that the synthetic texture of the CAD model may not be ideal for rendering and comparing operations. We showed that if the object is represented as an implicit (neural) representation in the form of Neural Radiance Field (NeRF), it exhibits a more realistic rendering of the actual scene and retains the crucial spatial features, which makes the comparison more versatile. We evaluated our NeRF implementation of the render-and-compare method on transparent datasets and found that it surpassed the current state-of-the-art results. 

\end{abstract}

\section{Introduction}
\label{sec:intro}

Given an object's 2D image, inferring its translation and rotation from the CAD model coordinate frame to the camera coordinate frame is a challenging task. Precise estimation of the 6D pose of an object is a prerequisite for a wide range of applications such as robotic manipulations \cite{deep_object_pose, grasping_RBGD, self_supervised_pose_robot}, augmented reality \cite{learing_less, pose_net} or autonomous driving \cite{autonomos_driving_app}, where objects need to be interacted with or manipulated in real-world environments. The goal of the 6D pose estimation task is to determine 3 translational and 3 rotational parameters of the object with respect to the camera from which the object is being registered.

Many methods for estimating the 6D pose of an object rely on the existence of a 3D CAD model of the target object \cite{labbe2022megapose, DPOD, li2018deepim, single_shot6dpose, foundationposewen2024, foundpose, bundlesdfwen2023, 3DPoseLite, few_shot_6f, OSOP}.
Estimation of an object's 6D pose becomes extremely challenging when dealing with transparent, reflective, or in general non-Lambertian surfaces, which depend on the particular view direction and the background, therefore such texture patterns cannot be expressed in a regular explicit way. The variety of 6D pose estimation methods \cite{foundationposewen2024, SSPPOSE, FFB6D, PVN3D, ove6d, SwinDePose} rely on depth information, which is considered to be a very strong prior knowledge of location in a 3D space. However, in the case of transparent objects, we cannot rely on this source of information because 3D sensors determine the object's position and shape based on the projection of light waves and its further reflection from the surface, and thus RGB-only methods are preferable.

\begin{figure}
     \centering
    \includegraphics[width=100mm]{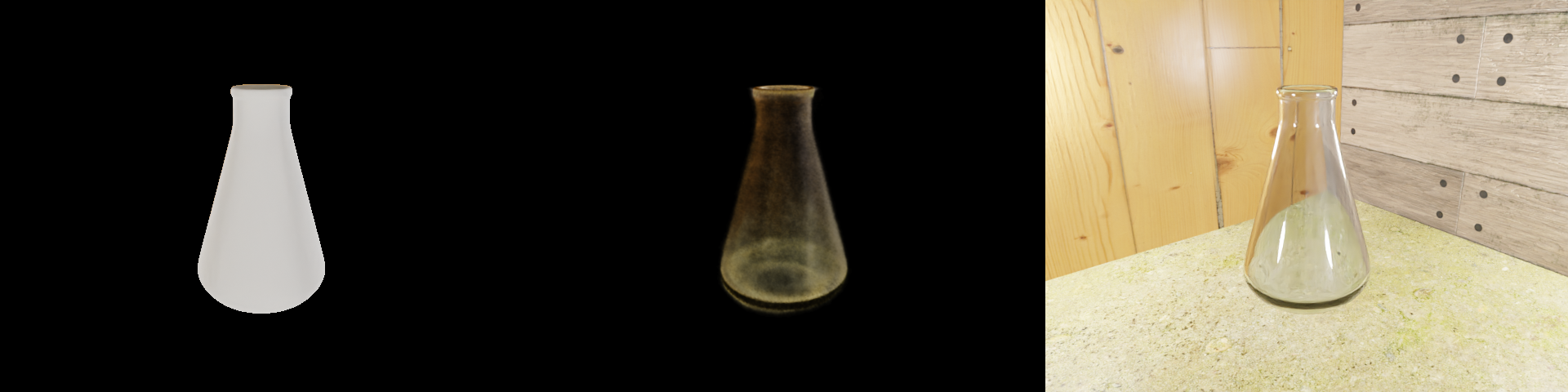}
      \caption{Illustration of the same object with different representations. On the left, there is a CAD model without any texture, in the middle there is the rendering of the same object using trained NeRF, and on the right there is the CAD model in the Blender scene. It can be seen there are big differences in their visual appearance.}
     \label{fig:img_mesh_nerf_blender}
 \end{figure}

We build our pipeline on top of the render-and-compare approach, which appears highly beneficial when dealing with textureless or non-opaque objects. Due to being similar to the template-based approach, render-and-compare relies more on high-level features such as shapes and contours than on low-level patterns. We utilize its robustness with respect to shading changes and featureless surfaces to perform 6D pose estimation of non-Lambertian objects. 
A variety of such methods \cite{OSOP, labbe2022megapose, li2018deepim, 3dobject_template} rely on template images, which are rendered from a given CAD model, by changing the camera position in order to capture different views of the object. However, representing non-opaque objects with CAD models is challenging, since transparency, translucency, or reflectance cannot be expressed in the form of predefined texture. The Neural Radiance Field (NeRF) \cite{mildenhall2020nerf} represents objects implicitly using a neural network optimized to fit the 3D scene. Unlike explicit representations such as triangular meshes, NeRF can exhibit varying radiance from different view directions, similar to the captured images. In Fig.~\ref{fig:img_mesh_nerf_blender}, one can see the comparison of various representations of objects. 
 
The major contributions of this paper are threefold:
\begin{itemize}
  \item We present the pipeline for 6D pose estimation of unseen transparent objects from a single RGB image and a sparse set of multiview images of objects with pose information.
  \item We combine the classic render-and-compare method by utilizing NeRF for view synthesis to render high-quality and view-dependent hypotheses of transparent objects. 
  \item The proposed method is tested on four large-scale datasets of transparent and reflective household objects in complicated environments. For evaluation, we use a variety of metrics such as MSPD, MSSD, ADD, ADD-S, translation, rotation error, and 3D IoU \cite{hodan2020bop}. 
  
\end{itemize}

\section{Related work}
\label{sec:sota}

\subsection{3D reconstruction}

3D reconstruction is a computer vision technique that involves creating a three-dimensional representation of a scene or object using a collection of two-dimen\-sional images from different viewpoints. The purpose of 3D reconstruction is to recover the spatial geometry and structure of the scene, allowing the creation of a 3D model that accurately represents the object or environment in real life.

Nonlearning-based Multi-view stereo methods follow the SfM (structure-from-motion) pipeline and are broadly classified as incremental SfM \cite{incrementel_sfm ,phototoursim, schonberger2016structure} and global SfM \cite{two_view_constraint,lie_alebraic, roboust_rotation_and_translation, non-sequential, global_motion, global_fusion, camera_pose_registraiton, camera_triplet, efficient_sfm_graph, hybridSfm} which provide the camera poses for MVS dense reconstruction. The Multi-view stereo starts with the structure-from-motion pipeline by extracting features from the images and then matching those features among the images to get camera poses. Next, it performs the triangulation to generate a 3D point of the feature, while incrementally adding images to reconstruction, and optimizing consistently by removing outliers and performing bundle adjustment.

The following works \cite{optimize_view_graph_for_sfm, graph_based_consistent, video_google, scalable_recognition, fast_accu_image_matching, multiview_steropsis, schonberger2016pixelwise, photometric_bundle_adj, rendering_albitary_illumination, 2d-3d_matching,SurfelMeshing, EandEmathcing, image_based_loc_active, SCRAMSAC} provide approaches to improve the pipeline by optimizing graphs, parallel computation, feature matching, surface refinement, etc. Throughout the decade, learning-based improvement was also introduced \cite{stereo_matching_CNN, matchnet, learned_mulit_patch, point_based_MVS_network, deep_mvs} to improve MVS reconstruction.

Currently, 3D reconstruction for transparent objects has been a hot topic and many methods try to solve it by specialized camera setup using position-normal \cite{Position-Normal}, with the application of heat \cite{heat3d} or leveraging a large-scale dataset with deep learning \cite{seethrough}.

The introduction of Neural Radiance Field (NeRF) \cite{mildenhall2020nerf} opened a new window to explore the long-standing problem. The algebraic methods for multiple view geometry showed interesting and accurate results for surface reconstruction but unfortunately failed to infer about properties of the scene as well as the object property itself. More advanced usage of volume rendering equations can lead to more spectacular information about objects as shown in \cite{munkberg2021nvdiffrec}. Moreover, the usage of implicit scene representation such as occupancy network, Sign Distance Function (SDF), or just an multi-layer perceptron (MLP), etc., providing a more detailed surface of the scene can be seen in the following recent works \cite{Atzmon_2020_CVPR,chibane2020ndf,convOcc,grid_rep,icml2020_2086,sitzmann2019siren,deeplocal,convOcc}. 

\subsection{6D pose estimation}

The earliest methods, so-called, template-based methods \cite{Payet2011FromCT, Ulrich2012CombiningSA, Muoz2016Fast6P} relied
on a template database generated from the 3D CAD model. Templates are distributed and sampled to cover different views. The target image is then compared with each template, the similarity score is estimated, and the closest template is considered as the best match.

As soon as a large number of datasets, CAD models, and tools for synthetic data generation became available, learning-based methods were used in computer vision applications. Some methods tried to directly predict rotation and translation parameters \cite{pose_cnn, Deep-6DPose}, while other methods regressed either sparse
\cite{single_shot6dpose, bb8} or dense \cite{DPOD, Wang2021GDRNetGD, Park2019Pix2PosePC, Li2019CDPNCD} keypoints of particular objects with further application of the Perspective-n-Point (PnP) algorithm to get the final 6D pose. 

The above-listed approaches can show good performance only for specific objects that were used in the training process and consequently will perform poorly if some novel unseen object appears. This drawback brought a new generation of algorithms that are able to predict the 6D pose of unseen objects. 
The set of methods \cite{NOCS, VI-Net, secondpose, AG-Pose, fs-net, GPV-Pose, Chen2021SGPASP, Lin2021DualPoseNetC6}, the so-called category level, is able to generalize across a specific category, does not rely on availability of the 3D CAD model, therefore effectively addressing intraclass shape and texture variations. Instead of predicting just 6 degrees of freedom for rotation and translation, this type of method additionally estimates an object's size in 3 axes. 
Other methods solve pose estimation by assuming that each object is a separate instance. 
Some of these methods are called model-free and do not require CAD models but use the sequence of images with annotated relative camera poses. For example, \cite{sun2022onepose, He2023OnePoseKO} leverage the SfM pipeline to reconstruct the sparse point cloud and perform 2D-3D matching, or \cite{liu2022gen6d} predicts the closest view from the reference images and then refines the pose of the object. However, most methods tend to use available 3D CAD models with texture, from which template views are rendered and compared to the target by applying 2D-2D feature similarity \cite{OSOP, foundpose, nguyen2024gigaPose} with further establishment of 2D-3D correspondences, or predicting similarity scores for each view \cite{labbe2022megapose, foundationposewen2024} refining the most similar one iteratively \cite{li2018deepim}.
\cite{burde2024comparativeevaluation3dreconstruction} shows the comprehensive evaluation of how the quality of mesh affects the pose estimation in render-and-compare methods.

\subsection{6D pose estimation of non-Lambertian objects}
As we mentioned in the Introduction section, 6D pose estimation of non-Lambert\-ian objects is more challenging and unique than that of opaque objects, therefore considerably less work was done in this domain. %

\cite{ChiXu} applied instance segmentation and extracted ROI, where the transparent object is located. After that combination of color features, depth information, surface normals recovery, and 2D positional encoding is processed by the adapted dense fusion model \cite{wang2019densefusion}. Rotation is estimated by predicting the closest rotation anchor and additional refining rotation, while the translation vector is directly regressed.

\cite{byambaa} uses RGB-only information and utilizes CNN in order to predict belief maps of possible 2D locations of an object's keypoints with further application of the PnP algorithm to get final rotation and translation estimate. The authors generate a synthetic dataset of transparent objects for model training. 

In \cite{ghostpose} the authors regress 2D locations of projected 3D bounding box vertices. Using a two-camera setup, the 3D points are reconstructed from predictions. 

\cite{transnet} is a category-level 6D pose estimation method, which utilizes depth completion and surface normal estimation. The recovered depth and normals channels are concatenated with ray direction and RGB image into a 10-channel tensor, which represents a generalized point cloud. This representation is passed through point-cloud embedding transformer \cite{poinformer}, on top of which 4 decoders estimate within-category scale, translation, and rotations in X and Z axes.

The authors of Dex-NeRF \cite{DexNeRF} place multiple light sources in the scene and leverage specular reflections, which appear as white pixels under particular viewing directions of the camera, while for the rest of the viewing angles, a transparent surface is captured. Then, method estimates an accurate depth map, which is further used for grasping.

It is important to mention that most of the above-stated methods were not tested on datasets with heavy occlusions or diverse backgrounds. Moreover, unlike all of them, the proposed pipeline does not require training of the pose estimator on particular object's instances.

\section{Methodology}\label{sec:Method}
In this section, we explain the proposed pipeline for 6D object pose estimation of novel objects with non-Lambertian surfaces. Such a type of surface cannot be represented as a simple textured mesh, therefore, we leverage the ability of NeRF to represent its view-dependent effects. After that, we integrate NeRF view synthesis into the deep render-and-compare pipeline. Finally, we fine-tune the 6D object pose estimator on a synthetic dataset of transparent and reflective objects. 

For the whole pipeline, we assume the availability of the object's CAD model with the defined coordinate frame, which is used for extraction of objects' NeRFs representation.
Furthermore,  during the inference, our pipeline requires a single RGB image with corresponding intrinsic camera parameters and 2D object detection. It is used for cropping the region of interest, which is used later in the render-and-compare section of the pipeline, and selecting the corresponding NeRF for the rendering. 

\begin{figure}
    \centering
    \includegraphics[width=115mm]{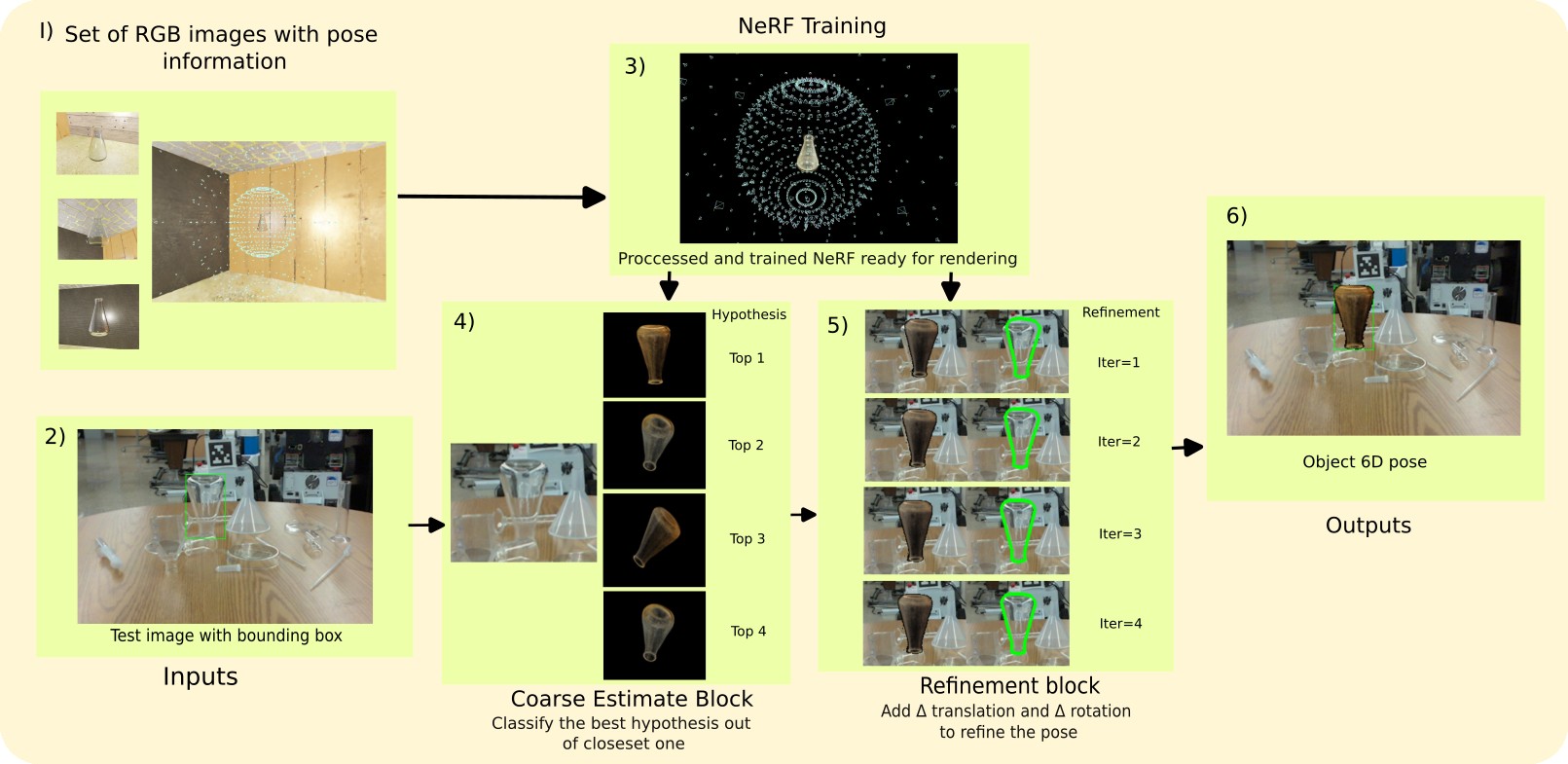}
    \caption{Working pipeline of the render-and-compare using NeRF as implicit representation. 1) Illustrate the camera poses and images rendered from the Blender scene which is fed to train the NeRF (model-free setup) 2) Shows the input of the Pipeline for inference, our method needs RGB image with the tight bounding box annotations for each object 3) NeRF is optimized for the scene and postprocessed to remove background noise 4) Describes the coarse estimation of the pose by performing classification task on the sampled rendered views 5) Illustration of refiner block which iteratively refines the pose by adding $\Delta t$ and $\Delta r$ to the pose 6) Finally, the resulting pose, in the figure one can see the pose of the NeRF overlayed on the exact pose of the object }
    \label{fig:pipeline}
\end{figure}

Apart from that, we operate on a model-free approach\footnote{https://bop.felk.cvut.cz/challenges/} from the BOP2024 challenge, i.e. a CAD model is not provided beforehand but instead a sparse set of multi-view images is given. The details of optimizing NeRF per each object for the experimentation setup are explained in Sec.~\ref{sec:Nerf data}. The process can be visualized from blocks 1) and 2) in Fig.~\ref{fig:pipeline}.

\subsection{Data collection}
\label{sec:Nerf data}

Given an object's CAD model, we apply reflection or transmission shaders in order to simulate realistic non-Lambertian properties. The object is placed in a cube with differently textured walls and four point lights near each wall. To acquire a rendered image, we define camera poses on the surfaces of 3 concentric spheres, while the object's coordinate frame remains fixed in the center. By dense sampling elevation, azimuth angles, and sphere radius, we cover a variety of views. This leads to capturing low- and high-frequency information. We fix the camera's intrinsic parameters and the inplane rotation of the camera throughout the whole process. After that, a set of high-quality rendered views with annotated ground truth camera poses is used to optimize the NeRF. After NeRF is optimized, we crop the scene using the 1.1 scale of the tight 3D bounding box calculated using the extension of meshes.

\subsection{ NeRF Traning}

NeRF infers density and point color inside the scene by optimizing an underlying continuous volumetric scene function using a sparse set of input views \cite{mildenhall2020nerf}. Using a volume rendering Eq.~(\ref{eq:volume rendering}), the NeRF parametrizes the volume density and color using the pixel color value. NeRF can render a novel photorealistic view of a scene with complicated geometry and appearance. 

\begin{equation}
\label{eq:volume rendering}
C(r) = \int_{t_{n}}^{t_f}T(t)\sigma(r(t)),d)dt 
\end{equation}
where,

\begin{equation}
    \label{eq:density function}
   T(t) = exp \left (- \int_{t_n}^{t} \sigma(r(s)) ds\right)   
\end{equation}

In Eq.~(\ref{eq:volume rendering}), $C$ is the color of the pixel, $T$ denotes the accumulated transmittance along the ray given by Eq.~(\ref{eq:density function}), $\sigma$ is the volume density, $r$ is the ray starting from o and traveling in direction d with magnitude~(\ref{eq:ray}).

\begin{equation}
    \label{eq:ray}
     r(t) = o + td
\end{equation}

\begin{figure}
    \centering
    \includegraphics[width=.8\textwidth]{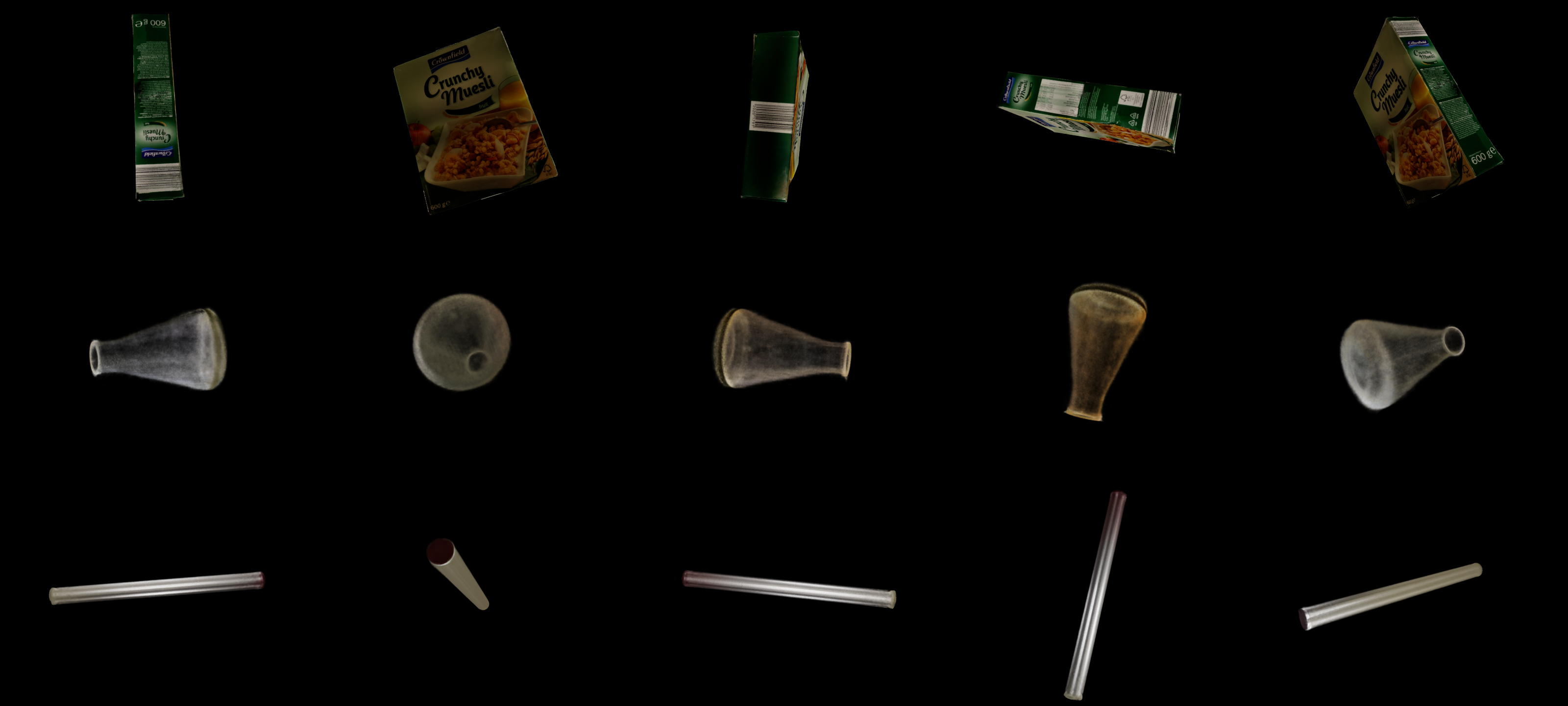}
    \caption{The following figure describes the visual appearance of various objects with the trained NeRF and their rendering from different views, the top row represents the textured box. The middle row is the rendering of transparent glass beakers, and at the bottom row, you can see the visualization of a metallic rod}
 \label{fig:img_multiview}
\end{figure}

A ray is shot through each pixel for novel view synthesis from a given pose. The neural network evaluates each point along a ray to predict occupancy and (view-dependent) color. Individual colors are accumulated on the basis of the predicted occupancies to obtain a final color for the pixel. The network is trained to reproduce the training images when using the view-synthesis approach described above. The NeRF can represent textured, transparent, and shiny objects with a view depending on the radiance, as can be seen in Fig. \ref{fig:img_multiview}.

Although the process of rendering a scene can be computationally expensive, various works have been done to improve the speed and rendering quality using Hash maps \cite{ngp} that allow for fast rendering capability. This allows for the pose estimation of an object in a few seconds, which makes it a reasonable choice for implementation.

\subsection{Pose Estimate using Coarse and Refinement block}

We used MegaPose6D as our base render-and-compare framework due to its highly optimized code. It uses ResNet34 \cite{He2015DeepRL} as the backbone architecture, which performs feature extraction and matching to the rendered image. We used the same initial guess of 1\,m with a random pose projected onto the test image. The depth is adjusted by fitting the projection into the input bounding box. For our implicit renderer, we pass the real-world scale to match the exact dimension of the object.

The coarse step renders 104 images of the classified object based on the input 2D bounding box and a randomly selected initial camera pose.
Then adjust the distance to the object based on the overlap of the bounding box and the projected object to the image space. 
After adjustment, the remaining 103 images are rendered based on the cube, which is defined by the center of the mesh and the adjusted initial camera pose.
This initial camera and the remaining 103 cameras are divided into quadruplets, each placed in corners, half-sides, and face centers.
The orientation of the quadruplet is designed to look at the center of the cube, while the quadruplets differ by in-plane rotation by 90\,°, 180\,°, and 270\,°.

We changed the explicit rendering with the Instant NGP renderer with the sample per pixel set to 4. This was chosen empirically to optimize the rendering speed and quality.
The rendered images are ranked on the basis of their similarity to the cropped inference image. 
This is done using the trained ResNet-34 backbone, which is very efficient in extracting and matching features from RGB images using the CNN network. 

The coarse estimate is passed on to the refiner block to further refine the pose to match the test image. The refiner step works similarly to the coarse step, except that the poses are sampled around the estimated coarse pose and refiner model provides update to rotation and translation. This step is performed iteratively a few times to more accurately match the pose of the object.

\subsection{Fine-tuning Procedure}

\begin{figure}
    \centering
    \includegraphics[width=0.8\linewidth]{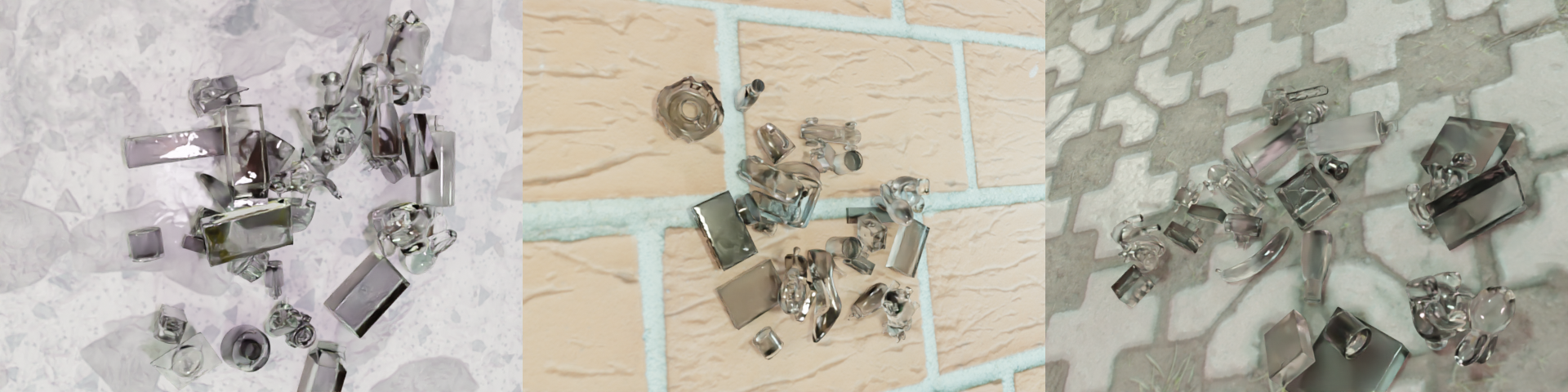}
    \caption{Dataset for finetuning of the pose estimator to improve the performance on transparent objects. The objects have been augmented with transparency shader. For each scene a random floor background is set with four sources of light around the objects}
    \label{fig:fine_tuning_dataset}
\end{figure}

The MegaPose6D was trained on a huge synthetic dataset of 2 million images and objects from GSO \cite{google_so} and ShapeNet \cite{shapenet2015}, we fine-tuned the MegaPose6D together with NeRF view synthesis on the dataset of 6000 images. To render this synthetic dataset of transparent objects, we use meshes from YCB-V \cite{pose_cnn}, HOPE \cite{tyree2022hope}, HomebrewedDB \cite{homebrew}, RU-APC \cite{ruapc}, and T-LESS \cite{tless} datasets and apply transparent shaders with slightly varying transmissivity and roughness to simulate different types of glass. Then we repeat the procedure for rendering the synthetic dataset, used by BOP \cite{bop_rendering}, by randomly sampling objects and using a physics simulator to make them fall in the plane. We obtain NeRF view synthesis by applying the same procedure as was described in the previous subsection. All synthetic data, seen in Fig. \ref{fig:fine_tuning_dataset}, were generated using BlenderProc\cite{bproc}. 

The coarse and refiner models of MegaPose6D \cite{labbe2022megapose} were trained each on 1 GPU for 500 000 iterations, which took 1 day. For the coarse estimator, the number of hypotheses was set to 5, 1 of which is always correct, and for the refiner model, the number of iterations was set to 3, while the learning rate was set to 1e-4 for both.  Online RGB image augmentation such as brightness, contrast, sharpness, and blur is applied during the fine-tuning procedure to prevent overfitting.

\subsection{Evaluation datasets}          
\label{sec:dataset}

\begin{figure}
  \centering
  \subfloat[][HouseCat6D\cite{jung2022housecat6d}]{\includegraphics[width=.3\textwidth]{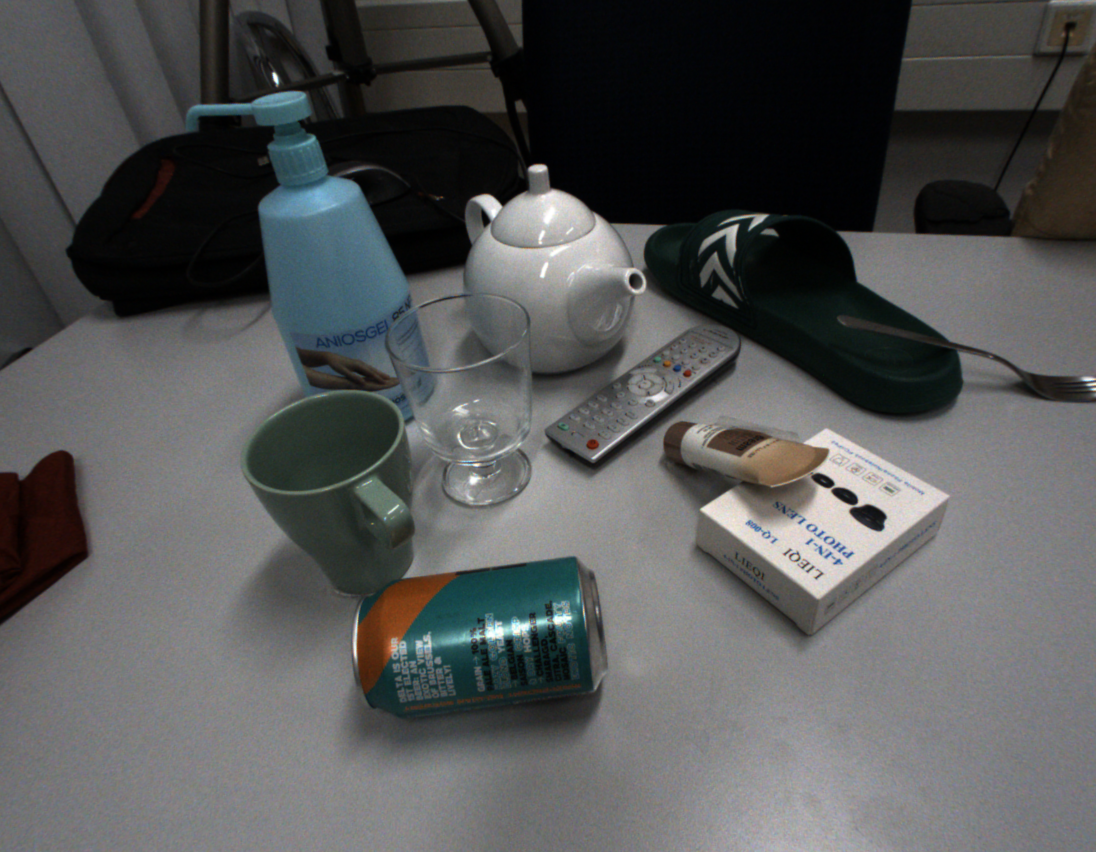}}\quad
  \subfloat[][ClearPose\cite{chen2022clearpose}]{\includegraphics[width=.3\textwidth]{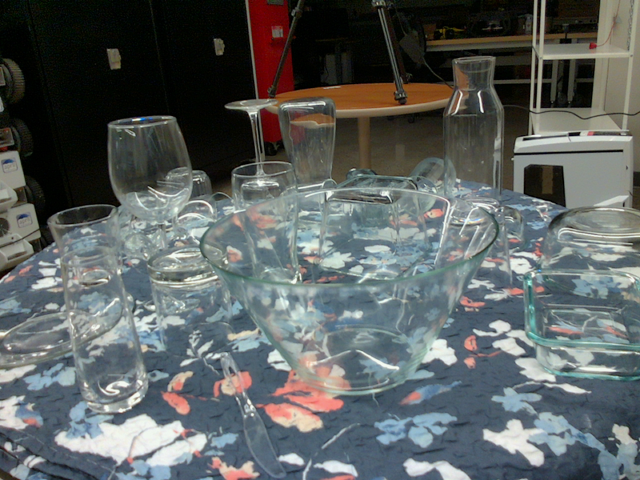}}\\
  \subfloat[][TransPose\cite{Jeonguyn2024TransPose}]{\includegraphics[width=.3\textwidth]{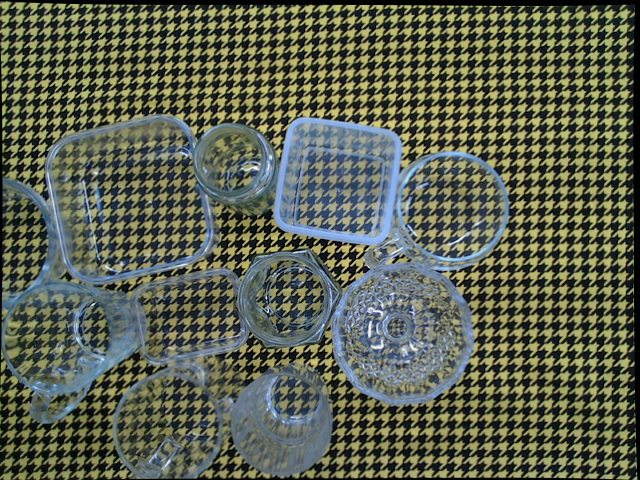}}\quad
  \subfloat[][DIMO\cite{roovere2022DIMO}]{\includegraphics[width=.3\textwidth]{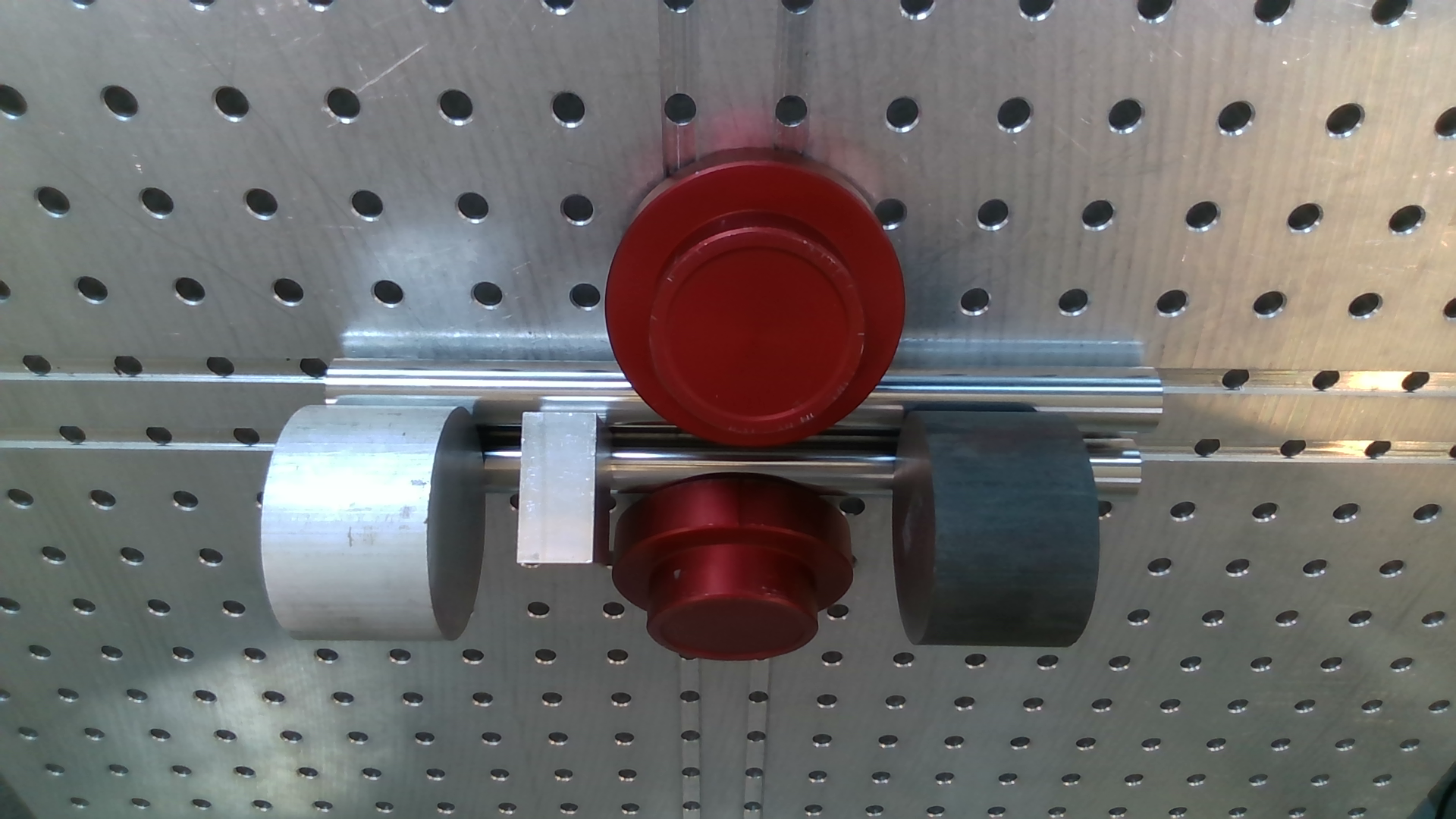}}
  \caption{Example images from the four benchmarked datasets. a) Test image from the HouseCat6D data set consisting of textured, shiny, metallic, and matte objects of different categories. b) The test image of the ClearPose dataset consists of glass utensils next to each other on the tabletop setup c) The TransPose test image consists of glassy and plastic objects with different optical properties cluttered on a tabletop setup d) The test image of DIMO dataset consists of colored, shiny, and matte finish metallic object on metallic surface}
  \label{fig:Datasets}
\end{figure}

Our results were evaluated on four datasets, HouseCat6D\cite{jung2022housecat6d}, Clearpose\cite{chen2022clearpose}, TRansPose\cite{Jeonguyn2024TransPose} and DIMO\cite{roovere2022DIMO}, out of which randomly selected images can be seen in Fig.~\ref{fig:Datasets}.

Similarly to HouseCat6D, ClearPose consists of 63 transparent or opaque household objects in various lighting conditions and occlusions.
We benchmark on the downsampled dataset\footnote{The downsampling is done by ClearPose authors by selecting every 100th image from each scene.} for a few reasons. Firstly, as the images were captured through continuous motion, there is a small difference in the poses of the object in two consecutive frames. Secondly, as our method does not require training or fine-tuning on specific objects, we present results on all the sets and objects. Finally, using the downsampled version decreases the inference time.

TRansPose, consisting of 99 transparent objects, is similar to ClearPose captured in continuous motion. Therefore, for the same reasons, we downsampled it by selecting every 10th image. The scenes were captured with a multi-camera setup consisting of RGB, RGBD, and TIR images.  As our method requires only RGB images and 2D detection as input, we relied only on RGB images and bounding boxes.
The DIMO dataset consists of six reflective metallic parts that exhibit a shiny appearance, which is possible to model through the NeRF view-dependent rendering.

All datasets were further converted to the standard BOP format in order to have a fair assessment of each data set and a comparative evaluation. The conversion scripts and the dataset can be shared if required\footnote{We are open to share these. If interested, write an email to the authors.}.

\subsection{Evaluation metrics}
We evaluated the results of our method on the BOP challenge error metrics, namely MSSD and MSPD introduced by Hodaň et al. \cite{hodan2020bop}.  
These metrics are strict due to the usage of maximal distance. Furthermore, they take into account the symmetries of each object.
The results are presented through AR$_{\mathrm{MSSD}}$ (Average Recall), 
AR$_{\mathrm{MSPD}}$, and their average AR$_{\mathrm{BOP}}$. For the calculation of AR$_{\mathrm{MSSD}}$ and AR$_{\mathrm{MSPD}}$ we followed the BOP challenge, which, for its calculation, uses thresholding based on the diameter of each of the objects. The thresholds range from 0 to 0.5 with a 0.05 step for the former and 0 pixels to 50 pixels with a step of 5 for the latter. The object diameter is defined as the maximal distance between any two points in the mesh. For AR$_{\mathrm{ADD}}$ and AR$_{\mathrm{ADI}}$ we use the thresholding of 0\,cm to 10\,cm with step 1\,cm. 

In addition, we conducted an additional evaluation of the dataset to compete against the leaderboard of the metrics provided. HouseCat6D is evaluated using 3DIoU and translation and rotation errors, which are model-free. ClearPose was evaluated on a model based on ADD and ADD(-S), metrics introduced by Hinterstoisser et al.\cite{Hinterstoisser2013ADDS}.

\section{Results}

\begin{figure}
  \centering
  \subfloat[][HouseCat6D]{\includegraphics[width=.3\textwidth]{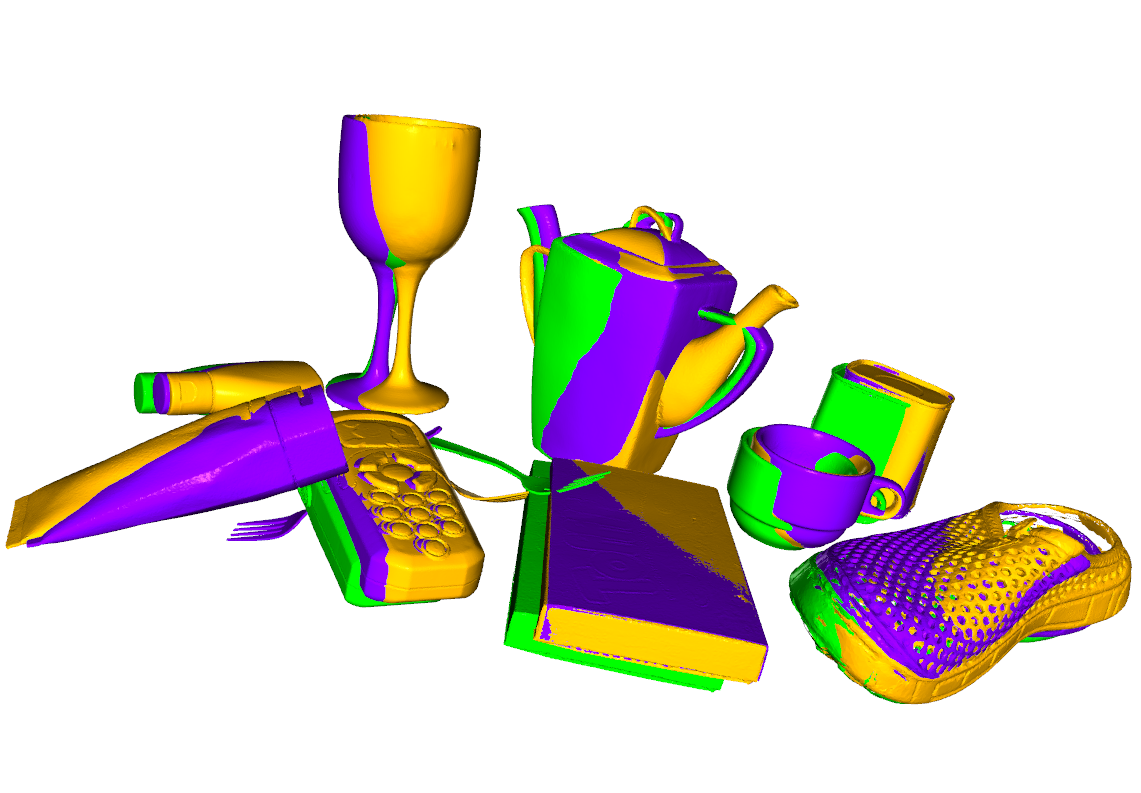}}\quad
  \subfloat[][ClearPose]{\includegraphics[width=.3\textwidth]{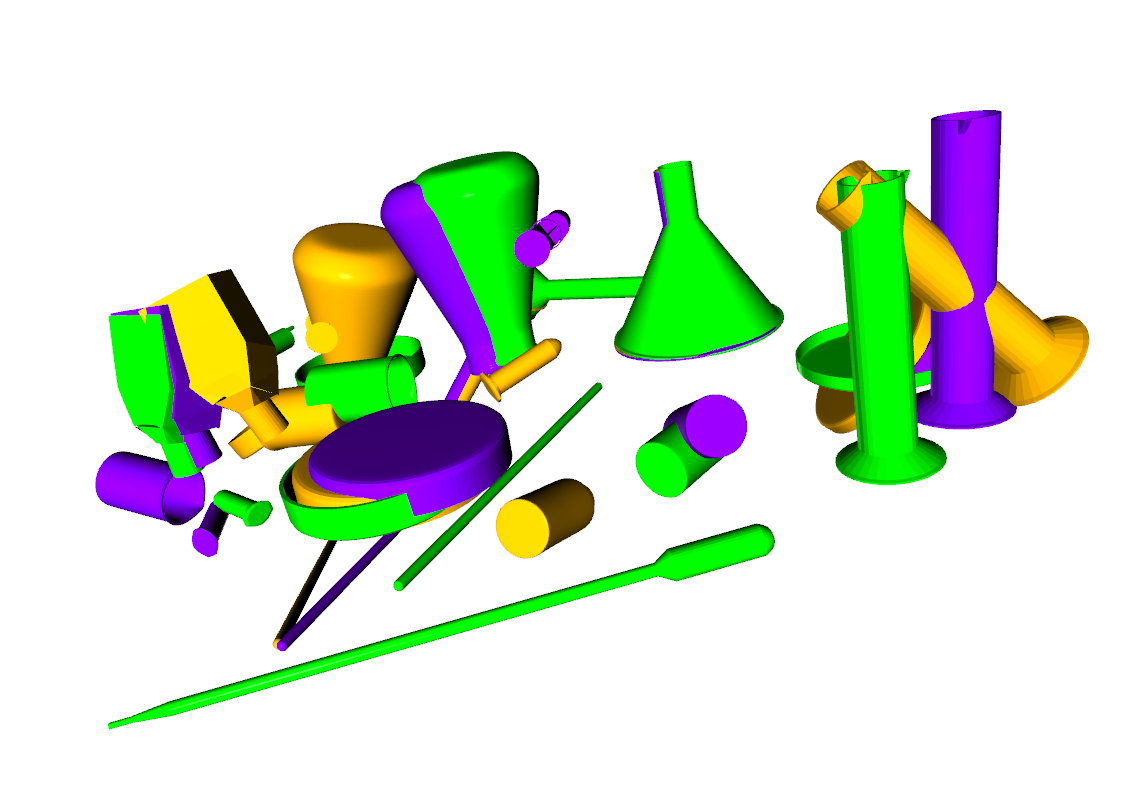}}\\
  \subfloat[][TransPose]{\includegraphics[width=.3\textwidth]{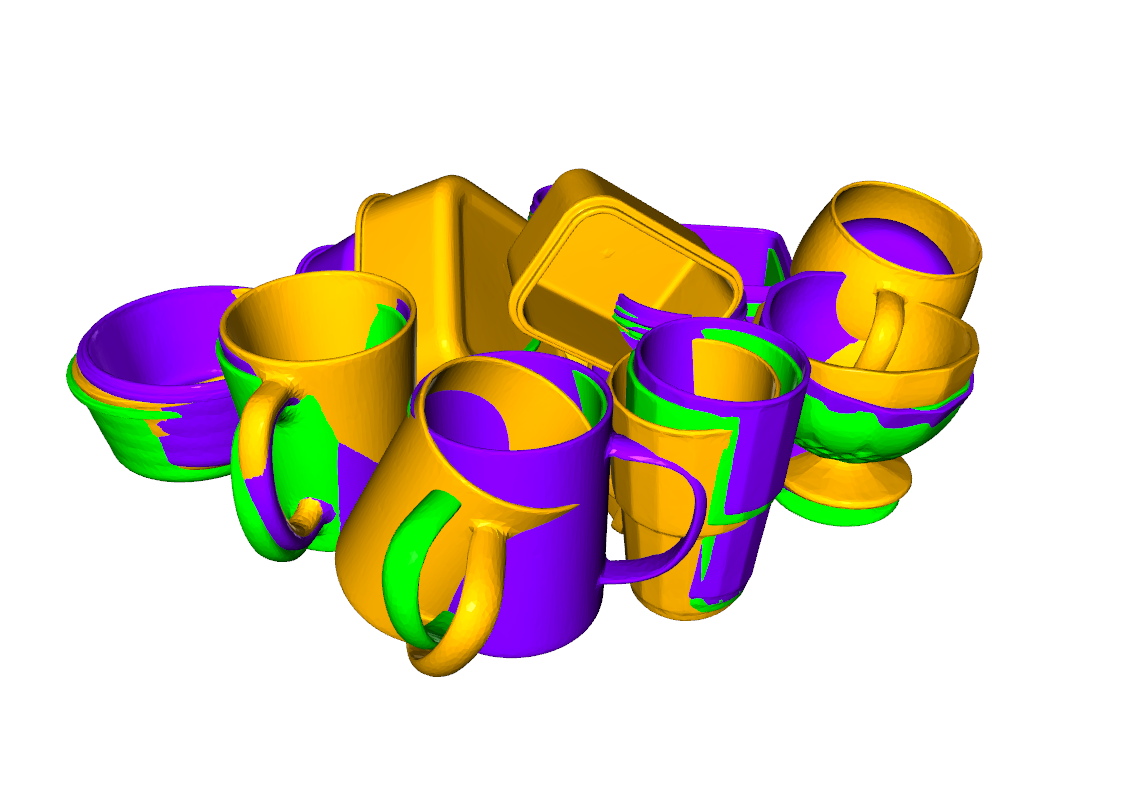}}\quad
  \subfloat[][DIMO]{\includegraphics[width=.3\textwidth]{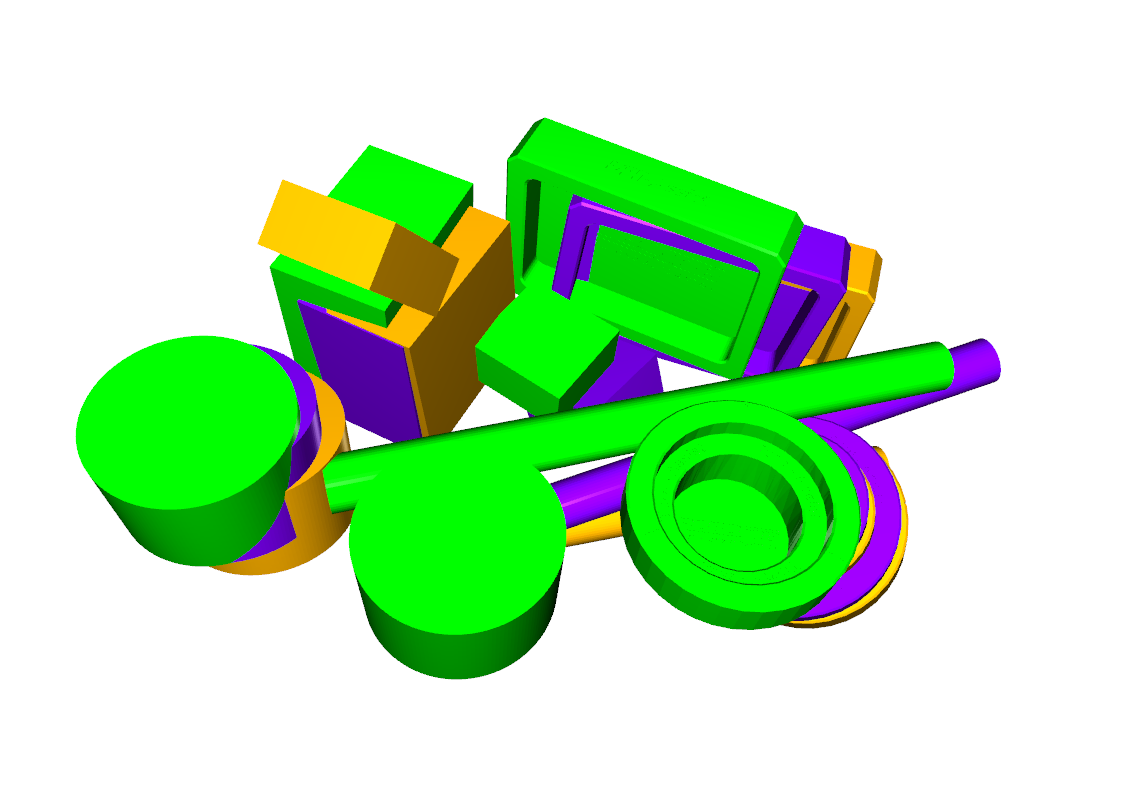}}
  \caption{Example of 3D visualization of the Ground-Truth(green) and predictions of our methods Ours(orange) and Ours(f)(purple). It can be seen on (a), that pose from the fine-tuned method Ours(f) overlapes much closer with the ground truth compared to Ours, which showes the effectivity of the fine-tuning procedure}
  \label{fig:3DPredictions}
\end{figure}

We present the result mainly with 3 methods:
\begin{enumerate}
    \item MegaPose6D - default rendering with meshes, which in case of transparent and reflective objects are textureless,
    \item Ours - MegaPose6D pipeline with replaced rendering by NeRFs, and
    \item Ours(f) - Finetuned MegaPose6D pipeline with NeRF rendering.
\end{enumerate}
Fig.~\ref{fig:3DPredictions} shows the visualization of randomly selected predictions for each dataset.

In~Tables~\ref{tab:HouseCat6D-Rete}~and~\ref{tab:HouseCat6D-3DIoU}  we present our results with a comparison against the method included in the leaderboard of the HouseCat6D dataset.  Our method outperforms the results previously reported. As this dataset consists of mostly Lambertian objects, we observe that the fine-tuning can lead to overall worse results; this was not surprising, as our fine-tuning consisted only of transparent objects. The highlight of our method can be seen in the case of the glass category, where the fine-tuning drastically increases the results, even with a stricter IoU threshold of 75.

\begin{table}
    \centering
     \caption{HouseCat6D rotation and translation errors. From the results, we can see that our method outperforms the other methods by a big margin. It can be seen that with the threshold of 10° 5cm, 72-78 percent of object poses fall in this range.}    \label{tab:HouseCat6D-Rete}
    \begin{adjustbox}{width=60mm}
    \begin{tabular}{c|cccc}
\thickerhline
Method & 5° 2cm & 5° 5cm & 10° 2cm & 10° 5cm\\
\thickerhline
FS-Net\cite{fs-net} & 3.3 & 4.2 & 17.1 & 21.6\\
GPV-Pose\cite{GPV-Pose} & 3.5 & 4.6 & 17.8 & 22.7\\
VI-Net\cite{VI-Net} & 8.4 & 10.3 & 20.5 & 29.1\\
AG-Pose\cite{AG-Pose} & 11.5 & 12.0 & 32.7 & 35.8\\
SecondPose\cite{secondpose} & 11.0 & 13.4 & 25.3 & 35.7\\
\hline
MegaPose6D & 54.0 & 54.0 & 60.7 & 60.7\\
Ours & \textbf{67.0} & \textbf{67.0} & \textbf{78.7} & \textbf{78.7}\\
Ours(f) & 61.2 & 61.2 & 72.7 & 72.7\\
\thickerhline
    \end{tabular}
\end{adjustbox}
\end{table}

\begin{table}
    \centering
    \caption{Evaluation of the test set of the HouseCat6D dataset on the 3DIoU with a threshold of 25, 50, and 75 percent. Each row shows the IoU scores on a different class of object, our method shows large score differences with the benchmarked method. Moreover, it can be seen that render-and-compare with NeRF and with finetuned pipeline shows a performance increase of 6 percent in IoU 25 and 14 percent on IoU 75 just with RGB image and without the need of being trained for a specific object set }
    \label{tab:HouseCat6D-3DIoU}
    \begin{adjustbox}{width=100mm}
    \begin{tabular}{c|c|ccccccccccc}
\thickerhline
\rotatebox[origin=c]{90}{ Metric } & Method &  Bottle & Box & Can & Cup & Remote & Teapot & Cutlery & Glass & Tube & Shoe & \textbf{All}\\
\thickerhline
{\multirow{9}{*}{\rotatebox[origin=c]{90}{IoU$_{25}$}}} &NOCS\cite{NOCS} &  41.9 & 43.3 & 81.9 & 68.8 & 81.8 & 24.3 & 14.7 & 95.4 & 21.0 & 26.4 & 50.0\\
&FS-Net\cite{fs-net} &   65.3 & 31.7 & 98.3 & 96.4 & 65.6 & 69.9 & 71.0 & 99.4 & 79.7 & 71.4 & 74.9\\
&GPV-Pose\cite{GPV-Pose}   & 66.8 & 31.4 & 98.6 & 96.7 & 65.7 & 75.4 & 70.9 & 99.6 & 76.9 & 67.4 & 74.9\\
&VI-Net\cite{VI-Net}   & 90.6 & 44.8 & 99.0 & 96.7 & 54.9 & 52.6 & 89.2 & 99.1 & 94.9 & 85.2 & 80.7\\
&AG-Pose\cite{AG-Pose}  & 82.3 & 57.2 & 97.1 & 97.9 & 87.0 & 63.4 & 77.2 & \textbf{100.0} & 83.4 & 72.0 & 81.8\\
&SecondPose\cite{secondpose}  & 94.5 & 54.5 & 98.5 & 99.8 & 53.6 & 81.0 & \textbf{93.5} & 99.3 & 75.6 & 86.9 & 83.7\\
\cline{2-13} 
&MegaPose6D   & 96.6 & 51.7 & 96.1 & 95.7 & 84.3 &\textbf{ 98.1} & 64.9 & 96.2 & 90.0 & 98.2 & 87.2\\
&Ours   & \textbf{99.8 }& \textbf{90.9} & \textbf{100.0} & \textbf{99.9} & \textbf{99.9 }& 97.3 & 62.0 & 96.9 & 99.5 & \textbf{99.5} & \textbf{94.6}\\
&Ours(f)   & \textbf{99.8} & 84.3 & 96.0 & 97.7 & 98.7 & 92.4 & 78.9 & 97.9 & \textbf{99.7 }& 98.9 & 94.4\\
\thickerhline
{\multirow{9}{*}{\rotatebox[origin=c]{90}{IoU$_{50}$}}} & NOCS \cite{NOCS} &  5.0 & 6.5 & 62.4 & 2.0 & 59.8 & 0.1 & 6.0 & 49.6 & 4.6 & 16.5 & 21.2\\
&FS-Net\cite{fs-net}  & 45.0 & 1.2 & 73.8 & 68.1 & 46.8 & 59.8 & 51.6 & 32.4 & 46.0 & 55.4 & 48.0\\
&GPV-Pose\cite{GPV-Pose}  & 45.6 & 1.1 & 75.2 & 69.0 & 46.9 & 61.6 & 52.0 & 62.7 & 42.4 & 50.2 & 50.7\\
&VI-Net\cite{GPV-Pose}   & 79.6 & 12.7 & 67.0 & 72.1 & 17.1 & 47.3 & \textbf{76.4 }& 93.7 & 36.0 & 62.4 & 56.4\\
&AG-Pose\cite{AG-Pose} & 62.8 & 7.7 & 83.6 & 79.6 & 66.2 & 60.9 & 62.0 & \textbf{99.4} & 53.4 & 50.0 & 62.5\\
&SecondPose\cite{secondpose}  & 79.8 & 23.7 & 93.2 & 82.9 & 35.4 & 71.0 & 74.4 & 92.5 & 35.6 & 73.0 & 66.1\\
\cline{2-13} 
&MegaPose6D  & 96.2 & 28.1 & 78.7 & 93.3 & 67.7 & \textbf{96.9} & 31.3 & 81.2 & 77.3 & 97.2 & 74.8\\
&Ours  & \textbf{99.4} & \textbf{90.4} & \textbf{99.7} & \textbf{99.3} & \textbf{96.3} & 96.8 & 25.6 & 85.0 & \textbf{99.0} & \textbf{97.7} & \textbf{88.9}\\
&Ours(f)  & 98.9 & 82.7 & 83.4 & 83.0 & 93.0 & 91.2 & 42.9 & 93.2 & 98.3 & 96.5 & 86.3\\
\thickerhline
{\multirow{3}{*}{\rotatebox[origin=c]{90}{IoU$_{75}$}}} & MegaPose6D  & 85.8 & 22.0 & \textbf{73.0} & \textbf{93.1} & \textbf{58.1} & \textbf{95.4} & \textbf{11.7} & 77.5 & 62.6 & \textbf{90.7} & \textbf{67.0}\\
& Ours  & \textbf{92.9} & \textbf{56.8} & 68.6 & 88.3 & 46.0 & 72.4 & 4.8 & 61.6 & \textbf{72.5} & 57.5 & 62.2\\
& Ours(f)  & 84.8 & 52.3 & 40.9 & 68.0 & 39.8 & 68.6 & 10.8 & \textbf{81.6 }& 67.7 & 50.9 & 56.5\\
\thickerhline
    \end{tabular}
    \end{adjustbox}
    
\end{table}

\begin{table}[]
    \centering
    \caption{Evaluation of the dataset is conducted using MSPD, MSSD, and their average recall. Additionally, we assess ADD and ADI metrics on the dataset separately. A more detailed evaluation of the clear pose for each category is provided in Table~\ref{tab:ClearPoseEval}}
    \label{tab:BOP-Eval}
    \begin{adjustbox}{width=80mm}
    \begin{tabular}{c|c|ccc|cc}

\thickerhline
Dataset & Method & AR$_{\mathrm{BOP}}$ & AR$_{\mathrm{MSPD}}$ & AR$_{\mathrm{MSSD}}$ & AR$_{\mathrm{ADD}}$ & AR$_{\mathrm{ADI}}$\\
\thickerhline
{\multirow{3}{*}{HouseCat6D}} & MegaPose6D & 66.76 & 70.98 & 62.54 & 73.12 & 91.50\\
 & Ours & \textbf{76.28} & \textbf{83.01} & \textbf{69.55} & \textbf{79.99} & 92.57\\
 & Ours(f) & 72.70 & 78.98 & 66.42 & 76.47 & \textbf{93.37}\\
\hline
{\multirow{3}{*}{ClearPose}} & MegaPose6D & 20.61 & 30.00 & 11.23 & 16.64 & 49.83\\
 & Ours & 26.90 & 38.13 & 15.67 & 19.61 & 53.20\\
 & Ours(f) & \textbf{28.85} & \textbf{40.02} & \textbf{17.68} & \textbf{20.38} & \textbf{54.13}\\
\hline
{\multirow{3}{*}{TRansPose}} & MegaPose6D & 20.55 & 21.92 & 19.19 & 25.20 & 59.33\\
 & Ours & 17.38 & 19.07 & 15.70 & 23.87 & 59.84\\
 & Ours(f) & \textbf{22.36} & \textbf{24.10} & \textbf{20.63} & \textbf{26.90} & \textbf{63.61}\\
\hline
{\multirow{3}{*}{DIMO}} & MegaPose6D & - & - & - & \textbf{23.56} & -\\
 & Ours & - & - & - & 20.33 & -\\
 & Ours(f) & - & - & - & 11.77 & -\\
\thickerhline

\end{tabular}
\end{adjustbox}
\end{table}

\begin{table}[]
    \centering
    \caption{Evaluation of 3DIoU rotation and translation errors on other datasets reveals that our method performs significantly worse on ClearPose, TransPose, and DIMO compared to the HouseCat6D dataset, as shown in Table~\ref{tab:HouseCat6D-Rete}. This reduced performance is likely due to the challenging nature of these datasets, which consist primarily of transparent or metallic objects. Transparent objects, in particular, create complex overlays, as illustrated in Fig.~\ref{fig:Datasets}, making accurate evaluation more difficult}
    \label{tab:3DIoUsRetes-alldatasets}
        \begin{adjustbox}{width=90mm}
    \begin{tabular}{c|c|ccccccc}
\thickerhline
Dataset & Method & IoU$_{25}$ & IoU$_{50}$ & IoU$_{75}$ & 5° 2cm & 5° 5cm & 10° 2cm & 10° 5cm\\
\thickerhline
{\multirow{3}{*}{ClearPose}} & MegaPose6D & 65.5 & 23.7 & 3.1 & 3.2 & 6.2 & 4.6 & 9.2\\
 & Ours & 69.0 & 28.1 & 4.1 & 4.0 & 8.1 & 6.3 & 13.1\\
 & Ours(f) & \textbf{69.6} & \textbf{29.0} & \textbf{5.1} & \textbf{5.1} & \textbf{11.2} &\textbf{ 6.9} & \textbf{15.8}\\
\hline
{\multirow{3}{*}{TRansPose}} & MegaPose6D & 59.0 & \textbf{20.5} & \textbf{7.0} & \textbf{9.7} & \textbf{10.9} & \textbf{12.6} & 14.9\\
 & Ours & 55.3 & 15.6 & 3.2 & 5.8 & 7.4 & 8.3 & 11.3\\
 & Ours(f) & \textbf{59.8} & 18.5 & 4.7 & 8.5 & 10.6 & 11.6 & \textbf{15.3}\\
\hline
{\multirow{3}{*}{DIMO}} & MegaPose6D & \textbf{63.7} & 14.8 & \textbf{3.9} & \textbf{4.1} & \textbf{4.5} & \textbf{5.2} & 5.9\\
 & Ours & 55.4 & 11.5 & 1.6 & 1.1 & 1.9 & 2.5 & 4.7\\
 & Ours(f) & 60.5 & \textbf{15.2} & 2.1 & 1.6 & 2.8 & 3.8 & \textbf{7.2}\\
\thickerhline

\end{tabular}
\end{adjustbox}
\end{table}

This correlates with the results presented in Tables~\ref{tab:BOP-Eval}~and~\ref{tab:3DIoUsRetes-alldatasets}, where it can be seen again that the fine-tuning increased the results for the datasets consisting of transparent objects, Clearpose and TRansPose. However, in the case of DIMO, we observed that fine-tuning on transparent objects, does not overcome the default MegaPose results, because our fine-tuning procedure was focused on transparent objects only.

\begin{table}[]
    \centering
     \caption{Result on selected scenes on ClearPose, compared with Xu et al. \cite{Xu2022ClearPoseTable} and FFB6D \cite{FFB6D} as presented in the paper. *Due to the large size of the dataset, we evaluate our results on the downsampled version, as mentioned in \ref{sec:dataset} to save some computing resources.
     }
    \label{tab:ClearPoseEval}
\begin{adjustbox}{width=\textwidth}
    \begin{tabular}{c|c|cccc|ccc}
\thickerhline
Testset & Metric & Xu et al. & FFB6D$_{r/r}$ & FFB6D$_{g/c}$ & FFB6D$_{g/g}$ & MegaPose6D* & Ours* & Ours(f)*\\
\thickerhline
\multirow{2}{*}{New Background} & Accuracy & 50.958 & 44.264 & 49.517 & \textbf{59.694} & 36.915 & 44.326 & 43.516\\
 & ADD(-S) & 45.233 & 43.452 & 47.691 & \textbf{58.224} & 41.802 & 45.991 & 46.243\\
\hline
\multirow{2}{*}{Heavy Occlusion} & Accuracy & 24.193 & 14.723 & 15.160 & 26.331 & 37.101 & 41.448 & \textbf{43.175}\\
 & ADD(-S) & 22.953 & 17.869 & 17.862 & 31.875 & 45.483 & 48.545 & \textbf{49.369}\\
\hline
\multirow{2}{*}{Translucent Cover} & Accuracy & 14.353 & 5.5617 & 4.5345 & 13.433 & 35.484 & 36.344 & \textbf{40.645}\\
 & ADD(-S) & 14.311 & 7.5983 & 5.8054 & 17.620 & 47.882 & 48.144 & \textbf{52.043}\\
\hline
\multirow{2}{*}{Opaque Distractor} & Accuracy & 42.630 & 0.4618 & 1.3331 & 2.3525 & 37.926 & 43.52 & \textbf{45.157}\\
 & ADD(-S) & 39.036 & 0.7628 & 1.5516 & 3.0685 & 46.011 & 47.849 & \textbf{48.816}\\
\hline
\multirow{2}{*}{Filled Liquid} & Accuracy & 34.500 & 7.6908 & 9.0584 & 16.228 & 44.782 & 51.21 & \textbf{51.955}\\
 & ADD(-S) & 32.251 & 11.153 & 10.828 & 18.583 & 53.729 & 57.959 & \textbf{59.555}\\
\hline
\multirow{2}{*}{Non Planar} & Accuracy & 21.024 & 7.4924 & 7.5843 & 15.567 & 35.244 & 42.038 & \textbf{43.312}\\
 & ADD(-S) & 18.411 & 7.8021 & 6.7339 & 16.986 & 41.197 & \textbf{46.117} & 44.301\\
\thickerhline
    \end{tabular}
\end{adjustbox}
\end{table}

\begin{table}[]
    \centering
    \caption{Comparison of requirements of methods. Our proposed solution does not require depth information, is not dependent on a particular category of objects and can be applied to novel unseen objects. The only requirement is an available CAD model}
    \label{tab:modality}
\begin{adjustbox}{width=100mm}

    \begin{tabular}{c|cccccccccc}
\thickerhline
Method & Ours & Xu et al. & FFB6D & FS-Net & GPV-Pose &VI-Net & AG-Pose &SecondPose \\
\thickerhline
Depth & \xmark  & \cmark& \cmark& \cmark& \cmark& \cmark& \cmark& \cmark\\ \hline
CAD model & \cmark  & \xmark& \xmark& \xmark& \xmark& \xmark& \xmark& \xmark\\ \hline
Category-dependent &\xmark  & \xmark& \xmark& \cmark& \cmark& \cmark& \cmark& \cmark\\ \hline
Novel objects &\cmark  & \xmark& \xmark& \cmark& \cmark& \cmark& \cmark& \cmark\\
\thickerhline

\end{tabular}
\end{adjustbox}
\end{table}

We compare our results to the other methods, as seen in table ~\ref{tab:ClearPoseEval}. One note due to the nature of the previous error calculation was only done from the downsampled dataset (i.e. every 100-th image) on the same scenes.
In Table~\ref{tab:modality}, different input requirements of the methods presented in this paper can be seen.

\section{Conclusion}

In this work, we present the render-and-compare using the NeRF which showed promising results of unseen objects with just RGB image. 
The NeRF as representation exhibits more photorealistic renderings when viewed from different view directions, allowing a better object pose estimation when using the render and compare method. In addition, we also showed the effect of fine-tuning the network with transparent objects, which showed promising results and provided an increase in accuracy specifically for transparent objects by a significant margin compared to the CAD model.

Our results are backed by our evaluation done with various challenging datasets which consist of not only transparent but also shiny metallic objects. Our results showed that our method is on par with most of the top-performing methods while just using a single RGB image. Due to the computationally intensive nature and long rendering times of NeRF, alternative representations such as Gaussian splatting can be used to enable faster rendering. This approach has potential applications in real-world robotics and should be considered for future work. Fine-tuning the weights with more variety of objects, in terms of transparent, translucent, and reflective properties of objects, or with Lambertian objects, could also lead to further generalization of the whole method.

\section{Acknowledgement}
This work contributes to the sustainability of project CZ.02.1.01/0.0/0.0/16\_ 026/0008432 Cluster 4.0 - Methodology of System Integration, financed by European Structural and Investment Funds and Operational Programme Research, Development and Education via Ministry of Education, Youth and Sports of the Czech Republic, and by the EU Horizon 2020 project RICAIP (grant agreement No. 857306).
This work was supported by the Ministry of Education, Youth and Sports of the Czech Republic through the e-INFRA CZ (ID:90254). 
This work was supported by the Grant Agency of the Czech Technical University in Prague, grant No.SGS23/172/OHK3/3T/13.
This work was co-funded by European Union under the project Robotics and advanced industrial production - ROBOPROX (reg. no. CZ.02.01.01/00/22\_008/0004590).

\bibliographystyle{splncs04}
\bibliography{main}
\end{document}